\documentclass[sigconf]{acmart}

\usepackage{algorithm}
\usepackage{algorithmic}
\usepackage{multirow,mathtools} 

\usepackage{xcolor,colortbl,pgf}
\usepackage{makecell}
\usepackage[para,online,flushleft]{threeparttable}

\usepackage[normalem]{ulem}

\newif\ifmodify
 \modifytrue 
\ifmodify

\newcommand{\cross}[1]{\textcolor{gray}{\sout{#1}}}
\else
\newcommand{\cross}[1]{}

\fi

\definecolor{Color0}{RGB}{0,0,252}
\definecolor{Color1}{HTML}{fff44f} 
\definecolor{Color2}{HTML}{d3d3d3} 
\definecolor{Color3}{HTML}{b9ffda}

\copyrightyear{2022} 
\acmYear{2022} 
\setcopyright{acmcopyright}\acmConference[ICCAD '22]{IEEE/ACM International
Conference on Computer-Aided Design}{}{San Diego,
CA, USA}
\acmBooktitle{IEEE/ACM International Conference on Computer-Aided Design (ICCAD
'22), October 30-November 3, 2022, San Diego, CA, USA}
\acmPrice{15.00}
\acmDOI{10.1145/3508352.3549379}
\acmISBN{978-1-4503-9217-4/22/10}

\begin{document}

\title{All-in-One: A Highly Representative DNN Pruning Framework   for  Edge Devices with Dynamic Power Management}

\author{
Yifan Gong$^{*1}$, Zheng Zhan$^{*1}$, Pu Zhao$^{1}$, Yushu Wu$^{1}$, Chao Wu$^{1}$, Caiwen Ding$^{2}$, Weiwen Jiang$^{3}$, Minghai Qin$^{1}$, Yanzhi Wang$^{1}$ 
}

\affiliation{$^*$Equal Contribution, $^1$Northeastern University, $^2$University of Connecticut, $^3$George Mason University }
\email{{gong.yifa,zhan.zhe,zhao.pu, yanz.wang}@northeastern.edu}

\renewcommand{\shortauthors}{Y. Gong and Z. Zhan, et al.}

\begin{abstract}
During the deployment of deep neural networks (DNNs) on edge devices, many research efforts are devoted to the limited hardware resource. However,  little attention is paid to the influence of dynamic power management. As edge devices typically only have a budget of energy with batteries (rather than almost unlimited  energy support on servers or workstations), their dynamic power management often changes the execution frequency as in the widely-used  dynamic voltage and frequency scaling (DVFS) technique. This leads to highly unstable inference speed performance, especially for computation-intensive DNN models, which can harm user experience and waste hardware resources. We firstly identify this problem and then propose All-in-One, a highly representative pruning framework to work with dynamic power management using DVFS. The framework can use only one set of model weights and soft masks (together with other auxiliary parameters of negligible storage) to represent multiple models of various pruning ratios. By re-configuring the model to the corresponding pruning ratio for a specific execution frequency (and voltage), we are able to achieve stable inference speed, i.e., keeping the difference in speed performance under various execution frequencies as small as possible. Our experiments demonstrate that our method not only achieves high accuracy for multiple models of different pruning ratios, but also reduces their variance of inference latency for various frequencies, with minimal memory consumption of only one model and one soft mask. 
\end{abstract}




\maketitle

\section{Introduction}

As deep neural networks (DNNs) can 
achieve  superior performance compared with traditional methods, 
they have been applied to a wide range of applications including classification \cite{he2016deep}, object detection \cite{bochkovskiy2020yolov4}, natural language processing \cite{vaswani2017attention}, and so on recently. Besides, due to the rapid increasing popularity of edge devices such as mobile phones, and tablets, there are ever-increasing demands for deploying DNNs on various resource-limited edge devices. 

When  deploying  DNNs from powerful servers (or workstations) to resource-intensive edge devices,  we need to deal with the  significant difference of (i) hardware resource and (ii) energy support. The hardware resource usually refers to the available memory, computation  units (such as multiplier–accumulator) and so on. Many works \cite{wen2016learning,guo2016dynamic,min20182pfpce,he2018amc,zhang2018systematic,zhang2018adam,zhang2021unified,jian2021radio,zhan2021achieving,wu2022compiler,yuan2021mest} investigate how to keep or improve the DNN performance under a more rigid resource constraint (such as limited memory or floating point operations per second (FLOPS)).

Though the resource limitation of edge devices receives much attention from researchers, there are little research efforts devoted to the energy support.   The  energy support refers to the power or energy (including the execution frequency and voltage)  to support the execution of DNNs when performing DNN inference on edge devices. Different from servers or workstations with plenty of power support, edge devices only have a budget of energy (such as battery) and need to adopt certain dynamic strategies to manage the energy usage. For example, if the battery level decreases to 15\% (or 10\%) on a mobile phone, it usually switches to the energy saving mode and reduces the execution frequency to achieve longer availability.  Besides energy saving mode,   the dynamic voltage and frequency scaling (DVFS) technique  is widely adopted on edge devices to adjust the power and speed settings   so that  the resource allotment for tasks can be optimized  and the power saving can be maximized.  In general, the  {current} energy support with dynamic frequencies is unstable  for the deployment of DNNs on edge devices.

The unstable energy support leads to unstable DNN performance in terms of inference speed or latency. If the execution frequency 
switches to a smaller value, the computation-intensive DNN models need more time to finish the computations, incurring   larger inference latency.  The unstable inference latency 
not only  harms    the user experience, but also wastes a lot of hardware resources.   {DNNs with original real-time inference capability} 
can hardly achieve real-time inference when the execution frequency decreases, incurring stuttering or lag, and poor user experience.  Besides, for safety-critical applications on real-time embedded systems like medical monitoring on smart watches, the deadline based scheduling is used 
to guarantee safety. The  deadlines are usually set up based on the worst-case execution time, which could waste many processor resources if the worst-case execution  time with low execution frequency is much larger than normal values.

To improve   user experience and save  hardware resource, it is necessary to make the DNN performance (especially speed performance) as stable as possible under different  levels of energy support (especially the execution frequency). Different from the hardware resource limitation  which   many research efforts focus on during DNN deployment, little attention is paid to the unstable energy support or dynamic frequency management.  Our work firstly identify the unstable energy and frequency problem during DNN deployment and propose a framework to deal with the limited hardware resource and unstable energy support. To deal with various execution frequencies, our framework can generate multiple sparse models with different sparsity ratios, such that  {higher-sparsity} 
models with less computations can run with  a  smaller frequency, leading to smaller  {variance of the} inference time.  To achieve this, we propose   parametric pruning together with switchable thresholds and batch normalization. Thus, after training one model and one soft mask, we can use switchable thresholds to convert the soft mask into different binary  hard  masks of various sparsity ratios to deal with different execution frequencies. Besides, we further adopt switchable batch normalization to improve the accuracy performance of each sparse model. Our memory  cost    is only one highly representative single model and  a   soft mask, which is much lower than the cost of training  and storing  multiple sparse models separately. As a result,  our work can deal with the limited hardware resource and unstable energy  support on  edge devices simultaneously.  We summarize our contributions as follows. 
\begin{itemize}
  \item We firstly identify the unstable energy support problem with dynamic frequencies for DNN deployment on edge devices and evaluate the influence of dynamic frequency and power for DNN inference. 
  \item  To deal with the limited hardware resource and unstable energy  support on  edge devices, we propose  parametric pruning together with switchable threshold and batch normalization to generate one single highly representative  model and soft mask, which can represent multiple sparse models to work with different execution frequency.
  \item Our experimental results demonstrate that we can use a minimal memory cost with one model and one rescaled soft mask, to obtain multiple models of various sparsity levels with state-of-the-art classification accuracy compared with baseline methods.  Moreover, by using a higher-sparsity model for a lower frequency,  we can keep the latency under various frequencies as close as possible and significantly reduce the variance of the inference latency. 
\end{itemize}


\section{Related work}

\begin{figure*} [t]
     \centering
     \includegraphics[width=0.72\linewidth]{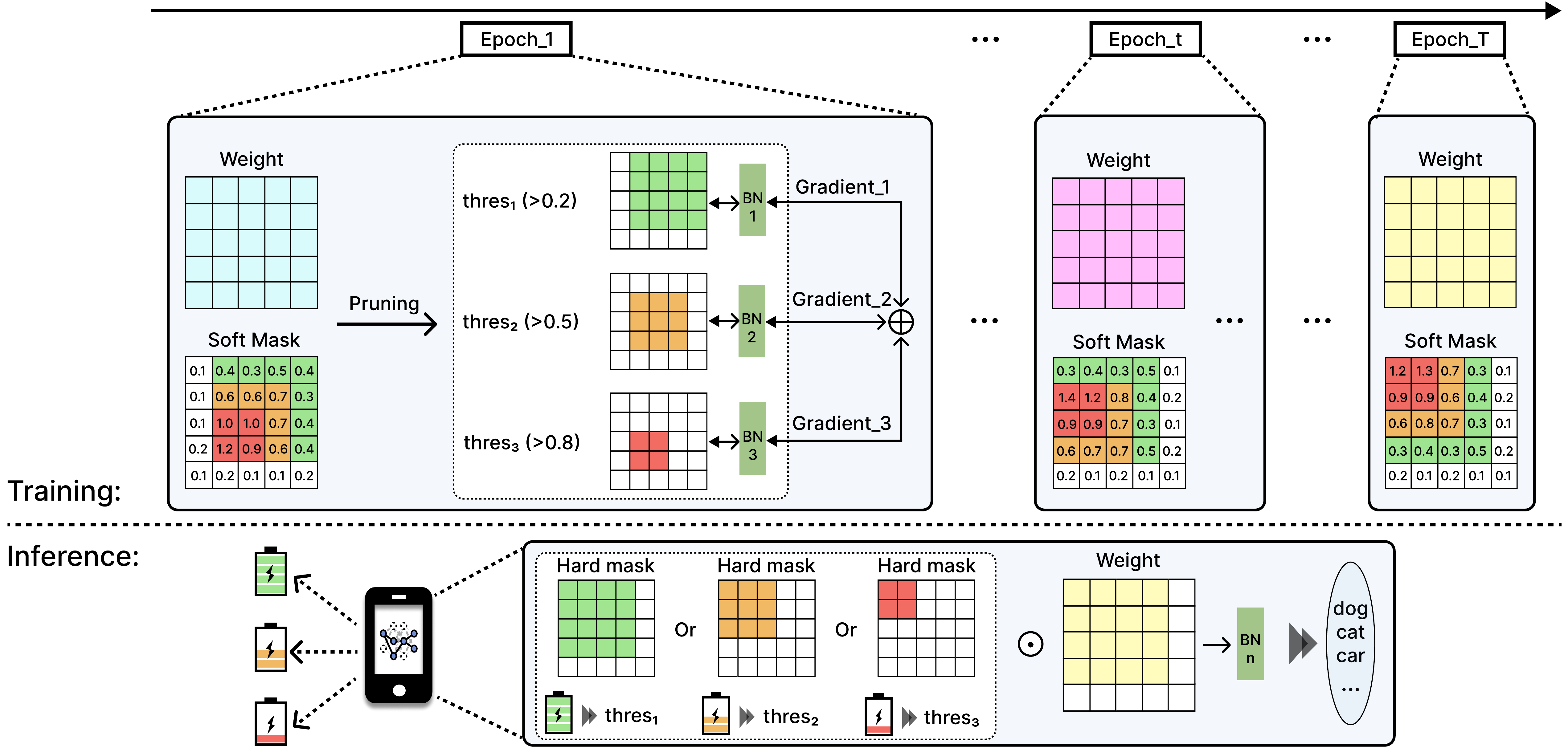}  
     \caption{Illustration of All-in-One framework in the case of three battery modes. Lower battery level requires a more sparse model to reduce the latency under a lower frequency. The framework learns \underline{one shared set of model weights} and \underline{one soft mask}. Each battery mode $n$ is paired with a threshold value $thres_n$ to transform the soft mask to a hard mask. The hard mask with a higher threshold value is the \underline{subset} of the hard mask with a lower threshold. During inference, there is no need to store the dense model, only \underline{the model weights in the largest compact model needs to be stored}. The model can be switched instantaneously according to $thres_n$ when the battery mode changes. }
    \label{fig:overall_flow}  
    \vspace{-0.1in}
\end{figure*}

\subsection{Sparse Models} Weight pruning ~\cite{wen2016learning,guo2016dynamic,min20182pfpce,he2018amc,he2019filter} is an effective method to reduce model redundancy. 
Various pruning methods with different sparsity types have been proposed in the literature. 
Irregular pruning \cite{han2015learning} achieves high accuracy but the reduction in parameters cannot be transformed into inference speedup. Structured sparsity \cite{min20182pfpce,zhuang2018discrimination,zhu2018ijcai,ma2019tiny,zhao2019variational,Liu2020Autocompress} such as column pruning and row pruning are hardware-friendly, but the coarse-grained pruning nature causes accuracy degradation. Recently proposed fine-grained structured pruning regularities including pattern-based pruning \cite{ma2020pconv,niu2020patdnn,gong2020privacy} and block-based pruning \cite{dong2020rtmobile,li2020ss,ma2020blk,gong2021automatic} preserve high accuracy while maintaining structures that can be exploited by compiler for hardware accelerations. However, most of the sparse model methods focus on the fixed resource scenarios without considering the hardware environment change. The resulting models have one fixed architecture and parameters that only satisfy one sparsity ratio.

\subsection{Re-configurable DNNs}
To obtain multiple models with different sparsity ratios, the most straightforward method is to train multiple models with different sizes using the sparse model methods. However, this causes huge computation burden during training and requires storing multiple models on the device. To decrease the computation and storage cost,  recent works consider to train efficient architectures that can be re-configured at runtime. NS \cite{yu2018slimmable} and US-Nets \cite{yu2019universally} train a model with multiple switches corresponding to different number of channels. This method simply 
retains the channels in the front of each layer, which neglects the different importance of each channel to the whole model performance. Furthermore, removing entire channels incurs severe accuracy loss. Joslim \cite{chin2021joslim} extends in this direction by further allowing the joint optimization of width
configurations and weights. But the method still relies on the coarse-grained structure pruning, incurring accuracy loss. Paper \cite{cai2019once} proposes to train a large dense model and queries sub-models upon requirement. However, the method has to store a dense model and needs additional weights to achieve model transformation.  LCS \cite{nunez2021lcs} learns a compressible subspace to obtain models with different sparsity ratio settings, but the accuracy degradation is non-negligible. RT$^{3}$ \cite{song2021dancing} discovers multiple pattern sets with different sparsity ratio settings to be switched during runtime. The drawback is that the pattern sets have to be stored individually, consuming a large amount of storage, and are too irregular to achieve efficient hardware accelerations.

\section{Motivation}


 \begin{table}[]
\caption{Frequency/Voltage levels on Adreno 650 GPU of Qualcomm Snapdragon 865 Chipset in OnePlus T8 platform
} \label{tab:freq_level}
\scalebox{0.8}{
\begin{tabular}{l|ccccc}
\toprule
          & N$_{1}$ & N$_{2}$ & N$_{3}$ & N$_{4}$ & N$_{5}$\\ \hline
\makecell{clk/freq (mHz)} & 305  & 400 & 442 & 525 & 587\\ \hline
Vol (mV)   & 0.47-0.73 & 0.52-0.79 & 0.55-0.84
&0.58-0.89

 &0.61-0.90
\\ \bottomrule
\end{tabular}}
\vspace{-0.12in}
\end{table}

Many deep learning (DL) based applications  are deployed on edge devices such as mobile phones. Although many research efforts are devoted to the limited hardware resource on edge devices, little attention is paid to the influence of unstable energy support or  dynamic frequency management.  As edge devices typically only have a budget
of energy with batteries (rather than almost unlimited energy support on servers or workstations), their dynamic power management
often changes the execution frequency, leading to highly unstable
inference speed performance, especially for computation-intensive
DNN models. Table \ref{tab:freq_level} demonstrates the available Frequency/Voltage levels  on Adreno 650 GPU in OnePlus T8 platform. For example, if the battery level of a OnePlus T8 phone is below 15\%, the phone switches to the energy saving mode and the frequency is reduced to 52\% of that of full battery.  The significant reduction on the frequency leads to  longer inference time for these computation-intensive DL  applications.  We observe that the dense ResNet-18 on ImageNet dataset takes  18.86 ms to inference on the mobile GPU. But after the mobile frequency reduces to 52\%, the inference  time extends to 30.74 ms,  62.99\% longer.  Besides the energy saving mode,  the DVFS technique is widely applied on edge devices to adjust the power and frequency  to optimize resource allotment for tasks and maximize power saving. It is common that the applications need to run under dynamic  execution frequencies.

If the frequency becomes smaller, the model   may be slower with longer inference time. The  DNNs with original real-time inference capability  may fail to achieve   real-time inference and cause stuttering or lag, leading to poor user experience.
Besides user experience, many  hardware resource can be wasted. For example,  for safety-critical applications like medical monitoring on smart watches, the deadline based scheduling is used by the real-time scheduler to guarantee safety. The  deadlines are usually set up  based on the worst-case  execution time,  which could waste many processor resources if it is much larger than normal values.

To guarantee user experience and save hardware resource,  we should keep consistent  performance of DL applications (especially speed performance), under various execution frequencies. In this paper, we focus on image classification application as an example. But our method is not only limited to image  classification.   
To keep consistent speed performance under different frequencies,  inspired by the model pruning methods to achieve on-mobile real-time inference, we propose to use one model with switchable  different pruning masks corresponding to various sparsity ratios. Thus, for higher frequency where the computation could be faster, we can adopt the mask with fewer pruned weights and more computations. For lower frequency where the computation could be slower, we can switch to the mask with more pruned weights and fewer computations.  In this way, we may keep the inference latency  under different execution frequency as close as possible.

We highlight that it is non-trivial to design such one model with multiple switchable   pruning masks corresponding to various sparsity ratios. One naive alternative method is to train multiple pruned models, each model corresponding to one single sparsity ratio. However,   DL models usually consume a lot of memory and the available memory on mobile devices is limited.  Thus, we choose the method with one single model and multiple masks for this model.  There are still several requirements for this method: (i) The memory cost of each mask should be significantly smaller than the whole model to save memory cost.   (ii) The accuracy under each mask should be as high as possible. We should keep consistent classification performance after switching to different masks.

Our setting is different from previous pruning work \cite{wen2016learning,guo2016dynamic,min20182pfpce,he2018amc,he2019filter} as most pruning methods only design one single pruning mask for one model given one sparsity ratio. Specifically, our work designs multiple pruning masks simultaneously for one model and each mask can achieve high classification accuracy under the corresponding sparsity ratio. Compared with methods with only a single mask design, our work has the following advantages: (i) The training cost is saved. After the training, we can obtain multiple sparse masks corresponding to multiple sparsity ratios simultaneously, while the single mask design methods need to run their algorithm  multiple times to obtain multiple masks, incurring much larger training cost. (ii)  The memory cost is saved. The single mask design methods usually need to fine-tune based on each mask to improve the final accuracy. Thus, each mask is not able to represent the pruned model and  each pruned  model is stored instead of each mask, 
incurring much larger memory cost. Moreover, we further adopt switchable threshold and batch normalization to derive one highly representative model and one soft mask so that multiple sparse models can be obtained from them. Thus the memory cost can be further saved.

\section{Proposed Method}

The framework overview of All-in-One in the case of three battery modes is illustrated in Figure \ref{fig:overall_flow}. Our objective is to design one model and multiple masks where each mask corresponds to one sparsity ratio under a certain battery mode. The model can be easily switched during inference. When the frequency becomes smaller (battery level is lower), the model can switch to a mask with a larger sparsity ratio and becomes more sparse with fewer computations. Thus though the computation may be slower with lower   frequency, the inference time with the more sparse model may not change greatly. More specifically, as in the Figure \ref{fig:overall_flow}, a sparse model with green mask is used when the battery level is high. When the battery level decreases, the framework switches to use the orange mask. As the battery is almost out of power, the mask is switched to the red one. The switch can be achieved instantly without any retraining.

To achieve this, we  propose the  parametric pruning  method to parameterize the pruning. The traditional pruning usually needs the in-differentiable  sorting operations, which are unfriendly for the  model training especially with multiple sparse masks. With the proposed parametric pruning, we do not need to incorporate sorting operations, improving pruning and training efficiency. As different sparsity ratios cause different mean and variance for the feature map, we further adopt   switchable batch normalization (BN) to improve the classification accuracy   under each  sparsity ratio.

Based on the parametric pruning, we use switchable threshold $thres_n$ to switch among masks of various sparsity ratios during training. When deployed for inference, the soft mask and switchable thresholds can be further re-scaled into low-bit format. Therefore, we do not need to store $N$ binary masks for $N$ sparsity ratios. Instead, we only need to store one re-scaled mask together with multiple thresholds (each threshold can determine one binary mask based on the re-scaled mask). The re-scaled threshold transforms the re-scaled soft mask into the corresponding hard mask when the battery mode changes. The hard mask with a higher threshold value is the subset of the hard mask with a lower threshold value. For instance, in Figure \ref{fig:overall_flow}, the mask in red color which corresponds to the low battery mode is the subset of the mask in green color. Furthermore, instead of storing the entire dense model on the mobile device, only the weights in the largest compact model needs to be stored, which significantly reduces the memory cost. Taking Figure \ref{fig:overall_flow} as an example, there is no need to store the entire 5$\times$5 weight matrix. Alternatively, the 4$\times$4 weight matrix is stored. Detailed designs are presented in the following sections.

\subsection{Pruning Schemes} 

\begin{figure} [t]
     \centering
     \includegraphics[width=0.82\linewidth]{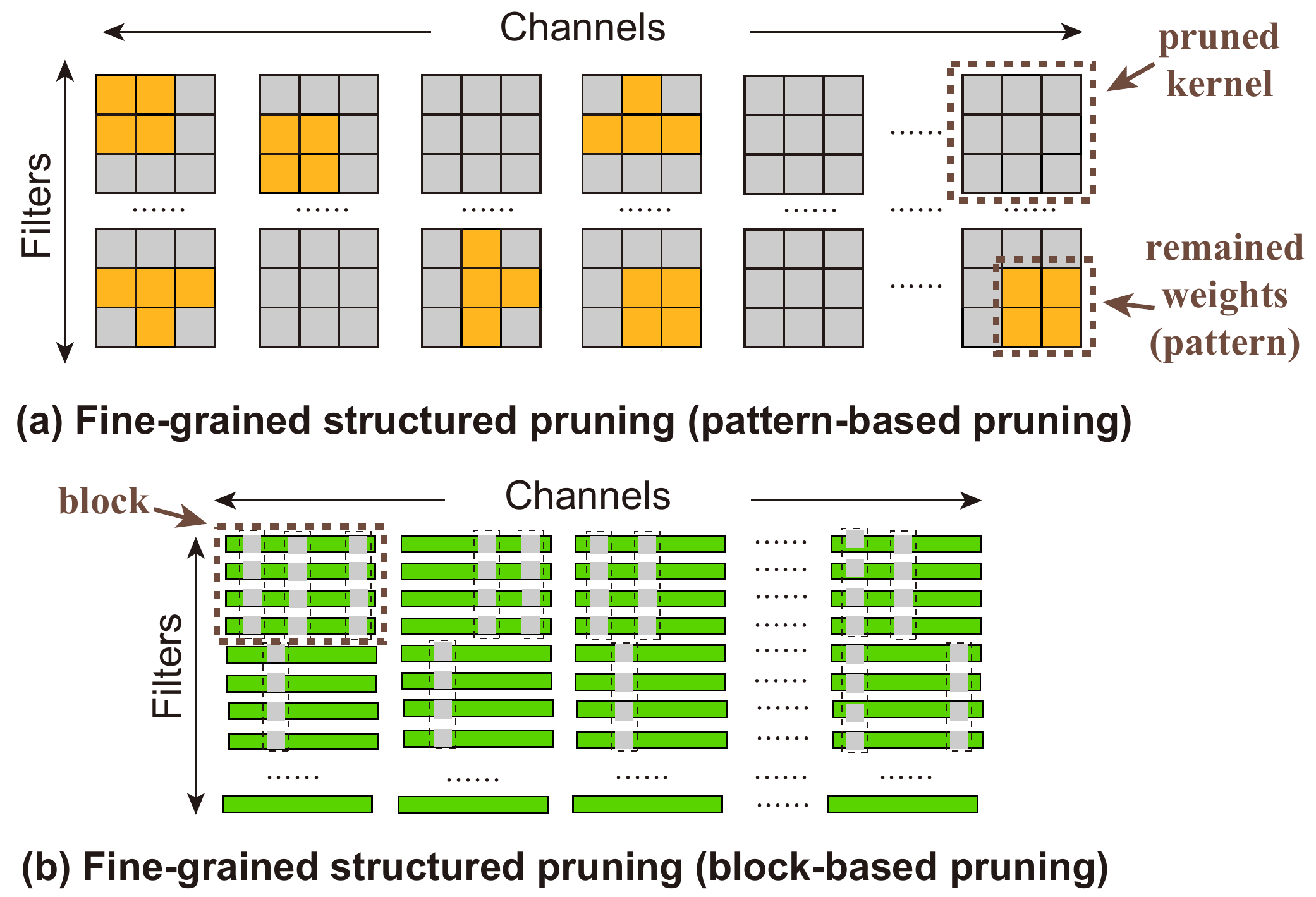}  
     \caption{Fine-grained structured pruning schemes. }
    \label{fig:pruning_scheme}  
    \vspace{-0.1in}
\end{figure}

In this work, we adopt fine-grained structured pruning including pattern-based pruning and block-based pruning, to efficiently accelerate the on-mobile inference while maintaining high accuracy. 
For both of the pruning schemes, the only set of weights to be stored is the largest compact model to accommodate the limited storage resources on the edge devices. 

\textbf{Pattern-based pruning.} 
Pattern-based pruning is a combination of kernel pattern pruning and connectivity pruning, as shown in Figure \ref{fig:pruning_scheme} (a), where grey color represents removed weight. Kernel pattern pruning removes a fixed number of weights in each kernel and the locations of the remaining weights form specific patterns. The total number of pattern styles in the pattern library is limited for hardware accelerations. In our framework, each kernel pattern reserves 4 non-zero weights to match the single-instruction multiple-data (SIMD) architecture of embedded CPU/GPU processors to maximize the hardware throughput. As the sparsity ratio is constant for kernel pattern pruning, connectivity pruning is adopted as the supplementary to kernel pattern pruning for a higher sparsity ratio. Connectivity pruning cuts the connections between certain input and output channels, which is equivalent as removing whole kernels. To provide a re-configurable sparse model with $N$ switches, our framework first applies kernel pattern pruning to obtain a compressed model. $N-1$ different levels of connectivity pruning are further applied to the kernel pattern pruned model so that $N$ models of different sparsity are available at runtime.  

\textbf{Block-based pruning.} 
Block-based pruning divides the weight matrix into  equal-sized blocks and apply independent column pruning for each block, as shown in Figure \ref{fig:pruning_scheme} (b). To find a re-configurable sparse model, our framework prunes columns in each block with $N$ different sparsity ratio settings to obtain well-trained $N$ switches.


\subsection{Parametric Pruning} \label{sec: parametric pruning}

We first need to make the pruning parametric to avoid the use of  in-differentiable  sorting operations. 
To achieve this, we assign importance scores to groups of weights. Let $\mathbf{w}_l \in \mathcal{R}^{O_l\times I_l\times H_l \times W_l}$ denote the weights for the $l$-th convolution (CONV) layer, with $O_l$ output channels, $I_l$ input channels, and kernels of size $H_l\times W_l$. The output feature of the $l$-th layer is represented as $\mathbf{a}_l \in \mathcal{R}^{B \times O_l \times f_l \times f_l'}$, with $O_l$ channels and $f_l \times f_l'$ feature size. The operation for the $l$-th layer is represented as $\mathbf{a}_l = \mathbf{w}_l \odot \mathbf{a}_{l-1}$, where $\odot$ denotes the convolution operation. For pattern-based pruning, as kernel pattern pruning results in a fixed sparsity ratio, a higher sparsity is achieved by pairing with connectivity pruning that removes whole kernels. Therefore, to perform connectivity pruning, for each kernel $\mathbf{w}^{(k)}_l\in \mathcal{R}^{H_l\times W_l}, k = 1,\cdots,O_lI_l$,  in the $l$-th layer,  we assign   an importance score $ s^{(k)}_l$  as a soft mask to weight the importance of each kernel. 
The importance scores in the $l$-th layer form the soft mask matrix $\mathbf{s}_l \in \mathcal{R}^{O_l\times I_l}$. For block-based pruning, the weight matrix $\mathbf{w}_l$ is first reshaped to a 2D matrix with size $O_l\times I_lH_lW_l$ and divided into $P_l$ equal-sized blocks with size $\mathcal{R}^{o_l\times i_lH_lW_l}$, namely, $\mathbf{w}_l = [\mathbf{w}_{l,1},\mathbf{w}_{l,2}, \cdots, \mathbf{w}_{l,P_l}]$. An importance score $ s^{(k)}_l$ is assigned for each column $k$ in every block $p$.

We use each element of $\mathbf s_l$  as the pruning indicator for corresponding weights. Larger value of $ s_l^{(k)}$ indicates a more important group of weights that should be preserved while smaller value means that the   weights may be removed. In this problem, we would like to achieve $N$ level sparsity, i.e., $N$  target sparsity ratios.  
The $n$-th target sparsity ratio   is paired with a predefined threshold value $thres_n$ to convert the score $ s_l^{(k)}$ into a binary mask as below,
{\small
\begin{equation} \label{eq:soft_to_hard}
  b_l^{(k)} = 
    \begin{cases}
    1, \ \  s_l^{(k)} \geq thres_n \\
    0, \ \   s_l^{(k)} < thres_n
    \end{cases}, 
\end{equation}}%
where $ b_l^{(k)} \in \{0, 1\}$ is the binarized $ s_l^{(k)}$. The non-binary  $\mathbf s_l$ is a soft mask and the binary  $\mathbf b_l$ is a hard mask. During the inference of model pruning and training, we first obtain the binary mask $\mathbf b_l$ from $ \mathbf s_l$. Then we apply the binary mask $ b_l^{(k)}$ to the corresponding weights  $\mathbf{w}^{(k)}_l$  following $ b_l^{(k)}  \mathbf{w}_l^{(k)}$ in each layer  
to achieve kernel or block pruning. $ b_l^{(k)}= 1 $ means that the group of weights is reserved and  $ b_l^{(k)}= 0 $ means the group of weights is pruned.


Thus we are able to obtain a binary mask for weights in each CONV layer. The next  problem  is how to make the soft mask trainable, as the binarization operation is  non-differentiable, leading to difficulties for back-propagation. To solve this,  we integrate Straight Through Estimator (STE) \cite{bengio2013setimating} as shown below, 
{\small
\begin{equation} \label{eq: binary}
\frac{\partial \mathcal L}{\partial  s_l^{(k)}} = \frac{\partial \mathcal L}{\partial  b_l^{(k)}},
\end{equation}}%
where we can directly pass the gradients through the binarization. STE is originally applied in quantization tasks \cite{liu2019learning,yin2019understanding} to avoid the non-differentiable issues.  
If we do not use STE, more complicated strategies may be applied to deal with  non-differentiable binary masks  such as \cite{guo2020dmcp, guan2020dais}.  
With binarization and STE, we are able to build a trainable soft mask to incorporate pruning in model training without using sorting operations.  
The trainable mask has the following advantages:  
(i) The soft mask can be efficiently trained along with the  network parameters via  gradient descent optimizers, thus   saving training cost  compared with \cite{guo2020dmcp, guan2020dais}. 
(ii) Different from previous methods \cite{han2015learning,yu2017compressing,he2017channel}, which determine the pruning  according to the  parameter magnitudes,  we  use the soft mask to serve as the pruning indicator, rather than the parameter magnitudes. Thus pruning is   decoupled  from the parameter magnitudes.

We can train the model weights and the soft masks simultaneously. As we need to achieve high accuracy for multiple sparsity ratios, one model and one soft mask is not able to achieve this. So we assign each sparsity ratio  a corresponding threshold and batch normalization parameters as introduced in Sec.~\ref{sec: thres} and Sec. \ref{sec: bn}, respectively. 
The classification loss can be expressed as $\mathcal L(\mathbf W, \mathbf S, thres_n, \mathbf B_n)$, where $\mathbf W$ denotes the   model weights, $\mathbf S$ represents the soft mask, $thres_n$ and $\mathbf B_n$ denote the threshold and  the BN parameters for the $n$-th sparsity ratio, respectively. 
Note that   $\mathbf W$ and   $\mathbf S$ are shared between different sparsity ratios when performing training and inference. Besides $\mathbf W$ and $\mathbf S$, each sparsity ratio has its specific threshold and BN parameters. 

\subsection{Switchable Threshold} \label{sec: thres}

\begin{figure} [t]
     \centering
     \includegraphics[width=0.92\linewidth]{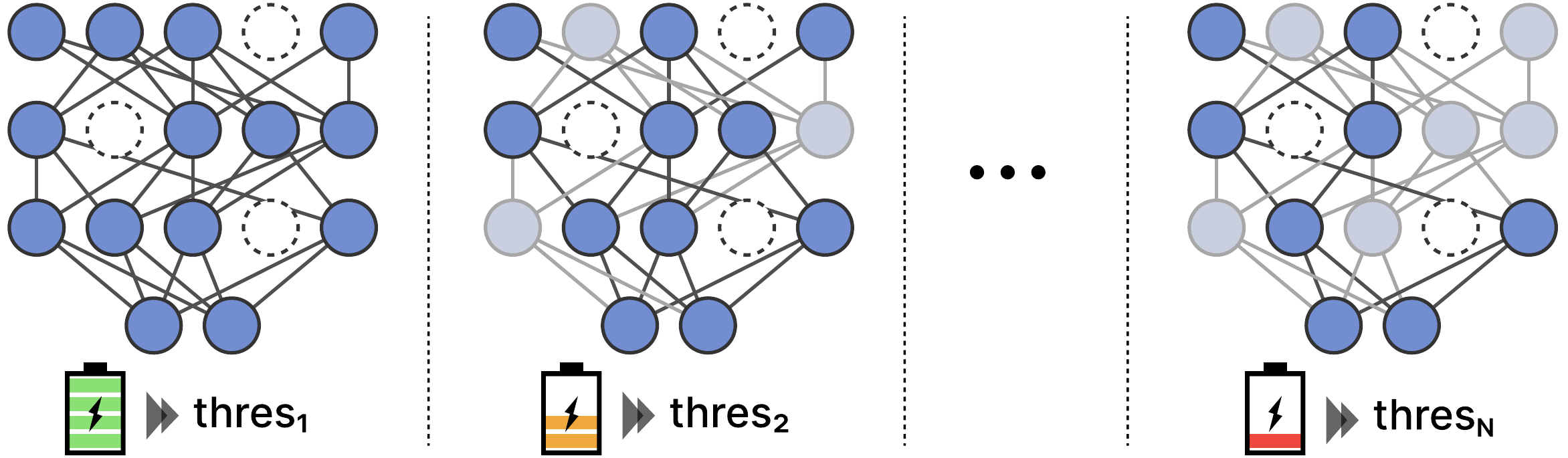}  
     \caption{Illustration of switchable threshold. Each battery mode corresponds to one threshold. Dark blue circles represent weights participating computation. White circles are weights that do not need to be stored. Light blue circles indicate the weights pruned by $thres_n$.  }
    \label{fig:switch_thres}  
\end{figure}


Given $N$ level sparsity or $N$ sparsity ratios, we can design $N$ masks where each mask can satisfy one sparsity ratio. However, multiple  masks still cost a lot of memory. To further reduce the memory cost, based on the parametric pruning, we can only use one soft mask with $N$ thresholds (i.e., $N$ scales with negligible memory cost). For each sparsity ratio, we can use the soft mask and the corresponding threshold  $thres_n$ following Eq.~(\ref{eq:soft_to_hard}) to obtain the binary mask for this sparsity ratio.  The illustration of the switchable threshold is shown in Figure. \ref{fig:switch_thres}. By using a higher threshold value, more weights are removed, which are denoted with the light blue circles. With the switchable threshold, the more sparse model uses a subset of the collection of weights in a more dense model.   


\subsection{Switchable Batch Normalization}  \label{sec: bn}

\begin{figure} [t]
     \centering
     \includegraphics[width=0.65\linewidth]{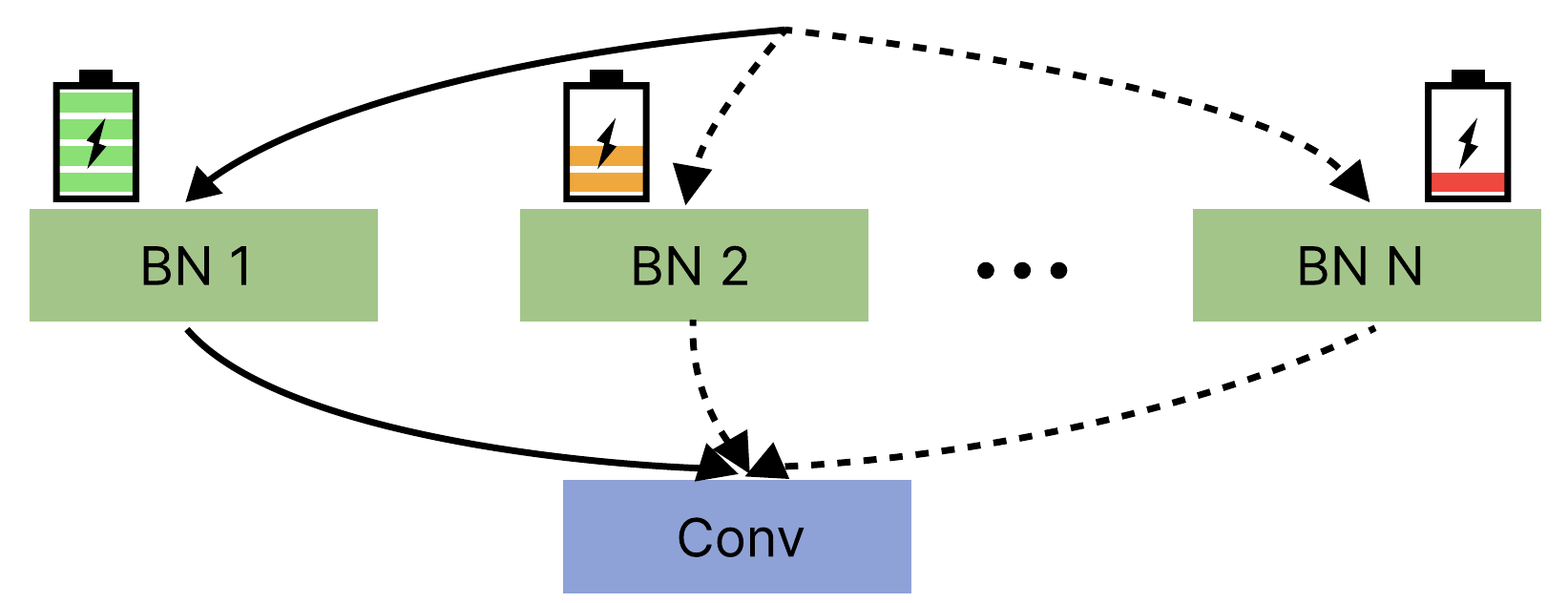}  
     \caption{Switchable Batch Normalization. }
    \label{fig:switch_bn}  
\end{figure}

As accumulating model parameters with different sparsity ratio results in different feature mean and variance, the discrepancy across different configurations leads to inaccurate statistics of shared  BN  layers. Therefore, to maintain high accuracy for each sparsity ratio, we create independent BN parameters for each sparsity ratio, as shown in Figure \ref{fig:switch_bn}. When inference under the sparsity ratio $n$, only the corresponding BN $n$ with parameters $\textbf{B}_n$ participates into the computation.  
With the switchable BN, for a model with $N$ target sparsity ratios, we   need to store $N$ set of BN parameters. 
In most cases, BN layers only have less than 1\% of the model size. Besides, the BN  runtime cost  is also negligible for deployment. Compared to separately trained models or using  two sets of weights for learning compressible subspaces \cite{nunez2021lcs}, the usage of switchable BN is memory efficient while keeping competitive accuracy performance. 

\begin{algorithm}[t]
\caption{Pruning with Switchable Batch Normalization}
\label{alg:switchable}
\begin{algorithmic}[1]
\REQUIRE Target MACs list $\mathbf{C} = \{\mathcal{C}_n\}_{n=1}^N$
\REQUIRE $\alpha$, $\beta$: step size hyperparameters
\STATE Randomly initialize score values $\mathbf{S}$
\STATE Initialize independent BN parameters for each target MACs 
\FOR{$i = 1, \cdots , n_{iters}$}
\STATE Get mini-batch of data $x$ and label $y$
  \FOR{target MACs $\mathcal{C}_n$}
 \STATE Switch to the threshold parameter $thres_n$ and the BN parameters $\mathbf B_n$ of current target MACs
 \STATE Compute loss $\mathcal L$ and sparse gradients
 \ENDFOR
 \STATE Apply accumulated gradient descent on weights
\ENDFOR
\end{algorithmic}
\end{algorithm}

\subsection{Training Method } \label{sec: training loss}

With the differentiable  soft mask, we can train and prune the model via SGD simultaneously with the  loss function,
{\small
\begin{equation}
    \min_{\mathbf W, \mathbf S, \mathbf B_n} \mathcal L(\mathbf W, \mathbf S, thres_n, \mathbf B_n) + \gamma \cdot \mathcal L_{reg}(\mathbf S, thres_n),
\end{equation}
}%
where $\mathcal L$ is the cross entropy classification loss, and   $\mathcal L_{reg}$ is the regularization term related to the computation complexity. For simplicity, we take Multiply-Accumulate operations (MACs) as the constraint/target rather than parameter number to estimate on-device execution cost. 
$\gamma$ can weight 
the loss and stabilize training. 
$\mathcal L_{reg}$ can be simply defined as $\ell_2$ norm between current MACs and target MACs $\mathcal C$. For connectivity pruning, $\mathcal L_{reg}$ is defined as   
{\small
\begin{equation}
    \mathcal L_{reg} = \left\| \frac{4}{9}\sum_l ( f_l f_l'    H_l W_l (O_l I_l-\|\mathbf{b}_l\|_0 )) - \mathcal{C}  \right\|_2,
\end{equation}}%
where $\mathbf b_l$ is obtained by converting the score $\mathbf s_l$ into binary values according to Eq. (\ref{eq:soft_to_hard}), and $\|\mathbf{b}_l\|_0$ indicates the number of remaining kernels in the $l$-th layer. The constant 4/9 is attributed to kernel pattern pruning that reserves four weights in each $3\times3$ kernel. 

For block-based column pruning, $\mathcal L_{reg}$ is defined as   
\begin{equation}
    \mathcal L_{reg} = \left\| \sum_l \sum_{p} (f_l f_l' o_l  (i_lH_lW_l-\|\mathbf{b}_{lp}\|_0)) - \mathcal{C} \right \|_2,
\end{equation}
where $\|\mathbf{b}_{lp}\|_0$ represents the number of the remaining columns in the block $p$ of the $l$-th layer. 

Given $N$ different battery modes, we could set $N$  different target MACs 
to satisfy different sparsity ratio requirements and form a list of target MACs as $\mathbf{C} = \{\mathcal{C}_n\}_{n=1}^N$. With parametric pruning and switchable threshold \& BN, we can perform our training following Algorithm~\ref{alg:switchable}. 
With $N$ target MACs, in every iteration, for each target MACs, we first switch to the corresponding $thres_n$ and BN parameters, and then compute the loss with the target MACs to obtain gradients through back-propagation. After we collect and accumulate the gradients of all target MACs, we update the weights.   Note that the model weights and the soft masks are updated with the  gradients accumulated through all target MACs since 
they are shared between various target MACs. But the BN parameters are updated with the gradients of the single corresponding target MAC, rather than the accumulated gradients, as the BN parameters are switched when moving to another target sparsity ratio.


\subsection{Cost Analysis}

To satisfy the a total of $N$ sparsity ratios simultaneously, we only need to store one model, one re-scaled soft mask, $N$ re-scaled thresholds, and $N$ sets of BN, where the memory cost of thresholds and BN are negligible. With a well-trained soft mask, there is no need to store its precise 
floating point values during inference. To save the storage cost, the soft mask can be transformed/re-scaled into the low-bit format (named the re-scaled soft mask).
For instance, for  $N=3$, the re-scaled soft mask is composed of values of 0, 1, and 2, with element 2 indicating the scores for the most compact model corresponding to the lowest frequency. Besides, since we perform kernel or block pruning and the soft mask corresponds to the number of kernels or columns/rows in the block rather than all parameters in the model, the memory cost of the soft mask is much smaller than the model weights. Our method can greatly reduce the memory cost compared with using $N$ models for $N$ sparsity ratios.

\section{Experiment results}
In this section, we compare All-in-One with state-of-the-art re-configurable DNN methods. The comparison is conducted on both CIFAR-10 and ImageNet datasets. 
We demonstrate the following: 
(i)  By comparing All-in-One with baseline methods, 
All-in-One can maintain high accuracy under different sparsity ratio settings with only one set of model weights. 
(ii) 
Through real implementations  on an off-the-shelf mobile device,  All-in-One can mitigate the problem of unstable inference time due to the change of battery level. 

\subsection{Experiment Setting}

In order to evaluate whether All-in-One can consistently attain efficient pruned models for tasks with different complexities, we test on two major image classification datasets, i.e., CIFAR-10 and ImageNet. For CIFAR-10, we experiment with Resnet-18, Resnet-20, and Resnet-32. For ImageNet dataset, we experiment on Resnet-18 and MobileNet-v2. We conduct our multi-model training method on an $8\times$ NVIDIA GTX 1080Ti GPU server using Pytorch. During pruning, 
The SGD optimizer is utilized 
with a learning rate of $ 1\times10^{-3}$. 

All the mobile results are measured on the GPU of an OnePlus 8T mobile phone, which has a Li-Po 4500 mAh battery. The phone itself is equipped with a Qualcomm Snapdragon 865 mobile platform which including a Qualcomm Kryo 585 Octa-core CPU and a Qualcomm Adreno 650 GPU. The same compiler techniques in \cite{niu2020patdnn} are applied to optimize the DNN execution.



\subsection{Accuracy Performance}

The  accuracy performance on the CIFAR-10 dataset is shown in Table \ref{table:cifar_sparse}. We run our method with pattern-based pruning and block-based pruning respectively and obtain two sets of accuracy.  
The results show that All-in-One can maintain high accuracy for all switches with only one set of model weights. It significantly outperforms all other state-of-the-art re-configurable DNN methods on all three models with non-negligible improvements (such as our accuracy  above 94\% v.s. 92\% or lower from baseline methods for Resnet-32). 
Besides, our method with pattern-based pruning  usually achieves higher accuracy than ours   with block-based pruning.

The accuracy   improvements of All-in-One are mainly attributed to two aspects. The first is the fine-grained structured pruning scheme that removes weight at a finer granularity compared to US-Net \cite{yu2019universally} and LCS-S \cite{nunez2021lcs} that employs coarse-grained structured pruning. The second is the separate BN for each configuration that preserves the distinct mean and variance with different sparsity ratios, which is not considered in RT$^{3}$ \cite{song2021dancing}.

We further   evaluate on the ImageNet dataset and the results are shown in Table \ref{table:imagenet_sparse}. For Resnet-18, both pattern-based pruning and block-based pruning are conducted. For MobileNet-v2, as the computations are mainly from 1$\times$1 CONV layers, which is not suitable for pattern-based pruning, the results using block-based pruning are presented. According to the results, All-in-One again provides the best accuracy performance on both models for each switch in the re-configurable model with non-negligible  improvements compared with baseline methods (such as our accuracy   above 67\% v.s. 65\% or below from baselines for the 350M MACs case).

In general, All-in-One possesses the ability to maintain high accuracy for each switch with only one set of weights. When the battery level changes, the model can be easily switched with negligible accuracy degradation compared to other re-configurable DNN methods, greatly improving user experience.

\begin{table}[t!]
\caption{Accuracy of All-in-One and state-of-the-art re-configurable DNN methods on CIFAR-10 dataset.
}
\label{table:cifar_sparse}
\scalebox{0.85}{
\begin{tabular}{l||c||ccc}
\toprule 
 \multirow{2}{*}{\textbf{Method}} & \multirow{2}{*}{\textbf{Pruning Scheme}} &  

\multicolumn{3}{c}{\textbf{MACs under Resnet-18}} \\
&  & 250M & 80M & 40M \\\hline
RT$^{3}$ \cite{song2021dancing} & Unstructured & 92.86 & 91.88 & 87.10 \\
  \rowcolor[gray]{.9}  All-in-One (Block) & Structured & 94.98 & 94.61 & 94.01 \\
 \rowcolor[gray]{.9} All-in-One (Pattern) & Structured & 95.19 &  94.59& 93.89 \\
 
\hline

&  &\multicolumn{3}{c}{\textbf{MACs under Resnet-20}} \\
&  & 20M & 11M & 7M \\\hline

 US-Net \cite{yu2019universally} & Structured & 90.77 & 89.13  & 87.10   \\
 Joslim \cite{chin2021joslim} & Structured & 90.99 & 89.43  & 87.43   \\
 LCS-S \cite{nunez2021lcs} & Structured & 86.56 & 83.13 & 78.75 \\
 LCS-U \cite{nunez2021lcs} & Unstructured & 89.58 & 87.51 & 85.29  \\
RT$^{3}$ \cite{song2021dancing} & Unstructured & 88.13  & 86.62 & 83.83 \\
  \rowcolor[gray]{.9} All-in-One (Block) & Structured & 91.83 & 91.23 & 90.65 \\
 \rowcolor[gray]{.9} All-in-One (Pattern) & Structured & 92.49 &  90.97 & 90.42 \\
 \midrule
 
&  &\multicolumn{3}{c}{\textbf{MACs under Resnet-32}} \\
&  & 33M & 20M & 13M \\\hline
 US-Net \cite{yu2019universally} & Structured & 92.06 & 90.56  & 89.83  \\
  Joslim \cite{chin2021joslim} & Structured& 92.26 & 90.70  & 89.82   \\
RT$^{3}$ \cite{song2021dancing}& Unstructured & 91.04 & 90.03 & 86.31 \\
  \rowcolor[gray]{.9} All-in-One (Block) & Structured & 94.79 & 94.55 & 94.05 \\
 \rowcolor[gray]{.9} All-in-One (Pattern) & Structured & 95.07 &  94.63 & 94.31 \\
 \midrule
\bottomrule
\end{tabular}
}
\vspace{-0.15in}
\end{table}

\begin{table}[t!]
\caption{Top-1 accuracy performance on ImageNet dataset.
}
\label{table:imagenet_sparse}
\scalebox{0.84}{
\begin{tabular}{l||c||ccc}
\toprule 
\multirow{2}{*}{\textbf{Method}} & \multirow{2}{*}{\begin{tabular}{@{}c@{}} \textbf{Pruning}  \textbf{Scheme} \end{tabular}} & \multicolumn{3}{c}{\textbf{MACs under Resnet-18}} \\
&  & 850M & 480M & 350M \\\hline
 US-Net \cite{yu2019universally} & Structured &66.51 & 63.42 & 61.49 \\
 Joslim \cite{chin2021joslim} & Structured & 68.49 & 64.51 & 61.82 \\
 LCS-S \cite{nunez2021lcs} & Structured & 57.61 & 53.70 & 51.30 \\
 LCS-U \cite{nunez2021lcs} & Unstructured  & 68.75 & 67.51  & 65.62\\
 RT$^{3}$ \cite{song2021dancing} & Unstructured & 66.72 & 65.39  & 63.87\\
  \rowcolor[gray]{.9} All-in-One (Block)  & Structured & 69.97& 68.28&67.02 \\
 \rowcolor[gray]{.9}All-in-One (Pattern) & Structured & 70.22
 & 69.28
& 67.87 \\
\hline
 & & \multicolumn{3}{c}{\textbf{MACs under MobileNet-v2}} \\
& &  210M & 170M & 150M \\\hline
 US-Net \cite{yu2019universally} & Structured & 69.71 & 68.19 & 67.59 \\
 Joslim \cite{chin2021joslim} & Structured  & 70.60 & 69.90 & 69.10 \\
 RT$^{3}$ \cite{song2021dancing} & Unstructured  & 68.12 & 66.97  & 65.65\\
  \rowcolor[gray]{.9} All-in-One (Block) & Structured & 70.79& 70.01 &69.25 \\
\bottomrule
\end{tabular}
}
\vspace{-0.15in}
\end{table}

\subsection{Speed Performance}

To show that All-in-One can reduce the inference variance when the frequency changes, we show the speed performance comparison  of  single model and our method. Table~\ref{tab: N2} and ~\ref{tab: N3} show the results for ImageNet with $N=2$ and $N=3$  under various  GPU frequencies on mobile devices using Resnet-18.  Single model refers to the case with one single model to work with various frequencies. 

\begin{table}[]
\caption{Speed performance of single model and All-in-One with $N=2$  under various GPU frequencies. 
}
\scalebox{0.755}{
\begin{tabular}{c|c|c|c|c|c}
\toprule
\multirow{2}{*}{ \textbf{Method}} & \multirow{2}{*}{ \makecell{ \textbf{MACs under} \\ \textbf{Resnet-18}}} & \multicolumn{2}{c|}{\makecell{\textbf{Latency under  various} \\ \textbf{GPU frequencies (ms)}}} &      \multirow{2}{*}{ \makecell{\textbf{Var. of} \\ \textbf{latency}}}  & \multirow{2}{*}{\makecell{\textbf{Reduct.} \\ \textbf{rate}} }   \\ \cline{3-4} 
        &        & 400mHz&525mHz&  &\\ \hline \hline
 Dense &    1820M        &      33.7 &    28.1    & 15.68& --  \\ \hline \hline
  \makecell{Single model\\(Block)} &    480M        &        25.7    &    \textbf{21.6}    & 8.41 & 187$\times$  \\ \hline
  \makecell{Single model\\(Block)} &    350M                      &       \textbf{21.9} &    18.6  &   5.45   & 121$\times$ \\ \hline
\rowcolor[gray]{.9}      \makecell{All-in-One\\(Block)} &   480M,  350M                    &       \textbf{21.9} &   \textbf{21.6} &  0.045   &  1$\times$ \\ \hline \hline
  \makecell{Single model\\(Pattern)} &    480M        &        25.9    &     \textbf{22.0}    & 7.61 & 381$\times$  \\ \hline
  \makecell{Single model\\(Pattern)} &    350M                      &       \textbf{22.2} &    18.9  &   5.45   & 272$\times$ \\ \hline
\rowcolor[gray]{.9}      \makecell{All-in-One\\(Pattern)} &   480M,  350M                    &       \textbf{22.2} &   \textbf{22.0} &  0.02   &  1$\times$ \\ \bottomrule
\end{tabular}\label{tab: N2}}
\vspace{-0.15in}
\end{table}

From Table \ref{tab: N2} and \ref{tab: N3} we can notice that our structured pruning scheme can achieve inference speedup compared to the dense model on both datasets. Though reducing the latency, using a single sparse model cannot provide stable inference when the frequency changes. Different from single model, our method can dynamically reconfigure to a suitable model once the frequency switches. For example, in the case of $N=3$, our method can generate three sparse models with  different MACs of 850M, 480M, and 350M on ImageNet. The more sparse model is used for the lower frequency. For example, in Table~\ref{tab: N3}, given three frequencies 305mHz, 442mHz, and 587mHz, the model with 850M MACs is applied for the 587mHz frequency, the  model with 480M MACs is used for the 442mHz, and the model with 350M MACs is adopted for the 305mHz.

Combining the results in Table~\ref{tab: N2} and ~\ref{tab: N3}, we have the following observations. (i) A more sparse model runs faster under the same   frequency (such as 29.1ms for the model with 850M MACs v.s. 23.5ms for the model with 480M MACs under 442mHz using block-based pruning in Table~\ref{tab: N3}). (ii) Besides, for the same model, a higher frequency leads to a faster inference speed (such as 21.9ms under 400mHz v.s. 18.6ms under 525mHz for the model with 480M MACs in Table~\ref{tab: N2}). The significant changes of the inference latency under various frequencies lead to large variance of inference latency (such as 29.64 for the single block-pruned model with 850M MACs in  Table~\ref{tab: N3}).


To deal with the different execution frequencies, our method uses   models of various sparsity levels so that a more sparse model with less computations can be adopted for a lower frequency, leading to a smaller change to the inference latency. 
As shown in the results, the latency  under difference frequencies are very close with our method (such as 21.9ms and 21.6ms under 400mHz and 525mHz, respectively, in Table~\ref{tab: N2}). All-in-One can significantly reduce the variance of the inference latency. For example, In the case of $N=2$ on ImageNet, our variance with pattern-based pruning is just 0.02, much lower than the single  pattern-based sparse model with 480M MACs (7.61) or the sparse model with 350M MACs (5.45), with a  reduction rate on the latency  variance as high as 381$\times$. In the case of $N=3$, our variance is 0.76 and 0.97 for block-based and pattern-based pruning, respectively, also significantly smaller than the variance of the single model method. 

\begin{figure} [t]
     \centering
     \includegraphics[width=0.8\linewidth]{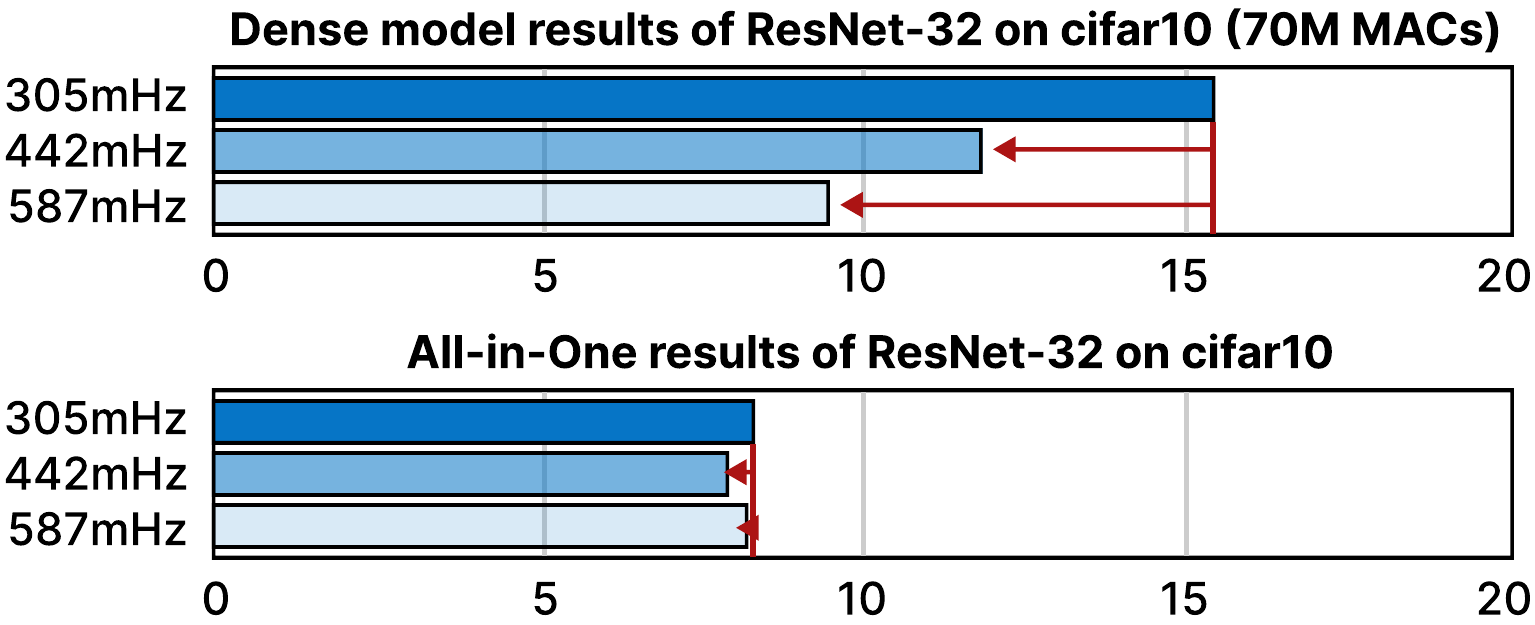}  
     \caption{Speed performance with $N=3$ of Resnet-32 on CIFAR-10 under various GPU frequencies.}
    \label{fig:res32_lat}  
    \vspace{-0.15in}
\end{figure}

We further show the variance on the CIFAR-10 dataset using Resnet-32 in Figure. \ref{fig:res32_lat}, where the red line is the worst-case execution time. All-in-One not only reduces the inference time compared to storing a dense model but also mitigates the inference variance problem due to frequency change.


\begin{table}[]

\caption{Speed performance of single model and All-in-One with $N=3$ under various GPU frequencies. 
}
\scalebox{0.73}{
\begin{tabular}{ c|c|c|c|c|c|c}
\toprule
\multirow{2}{*}{ \textbf{Method}} & \multirow{2}{*}{ \makecell{ \textbf{MACs under} \\ \textbf{Resnet-18}}} & \multicolumn{3}{c| }{ \makecell{ \textbf{Latency under various}\\ \textbf{GPU frequencies (ms)}}    }  & \multirow{2}{*}{\makecell{ \textbf{Var. of} \\  \textbf{latency}}}     & \multirow{2}{*}{ \makecell{\textbf{Reduct.}\\ \textbf{rate}}}                                       \\ \cline{3-5} 
 &    &            
 305mHz  &  442mHz  &  587mHz  &  & \\ \hline  \hline
    \makecell{Dense}   &    \makecell{1820M}                                              &  37.5     &  31.9         &    26.6   &  29.71  & --  \\ \hline  \hline
    \makecell{Single model \\ (Block)}   &    850M                                               &  35.7           &  29.1       &     \textbf{24.9}  &  29.64  & 39.0$\times$  \\ \hline
           \makecell{Single model \\ (Block)}   &              480M                                         &   29.7        & \textbf{23.5}   &    18.8  &  29.89  &  39.3$\times$ \\ \hline
            \makecell{Single model \\ (Block)}    &             350M                             &    \textbf{25.1}      &    20.3    &     15.8  & 21.63  &  28.5$\times$   \\ \hline
     \rowcolor[gray]{.9}    \makecell{All-in-One\\(Block)}   &     \makecell{850M, 480M,\\ 350M}         &    \textbf{25.1}      &    \textbf{23.5}    &     \textbf{24.9} &  0.76 & 1$\times$  \\ \hline \hline
         \makecell{Single model \\ (Pattern)}   &    850M                                               &  35.9          &  29.4          &     \textbf{25.5}  &  27.60  & 28.5$\times$  \\ \hline
           \makecell{Single model \\ (Pattern)}   &              480M                                         &   30.1       & \textbf{23.7}   &    19.0  &  31.04  &  32.0$\times$ \\ \hline
            \makecell{Single model \\ (Pattern)}    &             350M                             &    \textbf{25.3}      &    20.6    &     16.0  & 21.62  &  22.3$\times$   \\ \hline
     \rowcolor[gray]{.9}    \makecell{All-in-One\\(Pattern)}   &     \makecell{850M, 480M,\\ 350M}         &    \textbf{25.3}      &    \textbf{23.7}   &    \textbf{25.5} &  0.97 & 1$\times$  \\ \bottomrule 
\end{tabular}}\label{tab: N3}
\vspace{-0.15in}
\end{table}

\subsection{Memory Cost}

We compare the memory cost of our method and other baselines in Table~\ref{tab: memory_cost}. To obtain $N$ models of $N$ sparsity ratios, our method only need to store one model and one re-scaled soft mask. Since each element of the mask represents a group of weights, the memory cost of a soft mask is much smaller than that of a model. For other methods, the naive method to train multiple models each for a sparsity ratio needs to store $N$ models. 
RT$^3$  \cite{song2021dancing}  needs to store a model and $N$ masks so that each mask can convert the dense model into a sparse model corresponding to a specific sparsity ratio. LCS-U or LCS-S \cite{nunez2021lcs}  need to one model for the compressible points case, or two models for the compressible lines case, together with a mask. 
Some other methods may have similar memory cost with ours such as Joslim and US-Net. But as shown in Table~\ref{table:cifar_sparse} and \ref{table:imagenet_sparse}, unlike our method with high accuracy performance,  they are not able to achieve state-of-the-art accuracy on CIFAR-10 or ImageNet. More specifically, take storing a Resnet-18 model on ImageNet with $N=3$ sparsity ratios as an example, All-in-One only requires two additional sets of BNs, which accounts for only 0.14\% of the total number of parameters, and one set of re-scaled soft mask which accounts for 0.69\% of the total number of parameters. The total extra parameters incurred by All-in-One is 0.83\%. As All-in-One only needs to store the largest compact model, which is already compressed by 55\% of the total parameters, the overall memory cost is saved by 54.17\% comparing to saving a single dense Resnet-18. Therefore, All-in-One is memory-efficient.     


\begin{table}[]
\caption{Memory cost of various methods.
}
\scalebox{0.78}{
\begin{tabular}{c|c|c}
\toprule
\textbf{Method} & \textbf{Stored model num.} & \textbf{Stored mask num.} \\ \hline
 Multiple models      &    N    &  -   \\ \hline
  US-Net \cite{yu2019universally}      &          1           &      -              \\ \hline
  Joslim \cite{chin2021joslim}      &          1           &      1              \\ \hline  
  LCS-S  \cite{nunez2021lcs}      &          1 or 2           &      1              \\ \hline
  LCS-U \cite{nunez2021lcs}     &         1 or 2            &            1        \\ \hline
 RT$^3$  \cite{song2021dancing}    &        1            &          N          \\ \hline
\rowcolor[gray]{.9}  All-in-One     &            1         &          1          \\ \bottomrule
\end{tabular} \label{tab: memory_cost}}
\vspace{-0.1in}
\end{table}

\subsection{Ablation Study}

\begin{figure} [t]
     \centering
     \includegraphics[width=0.65\linewidth]{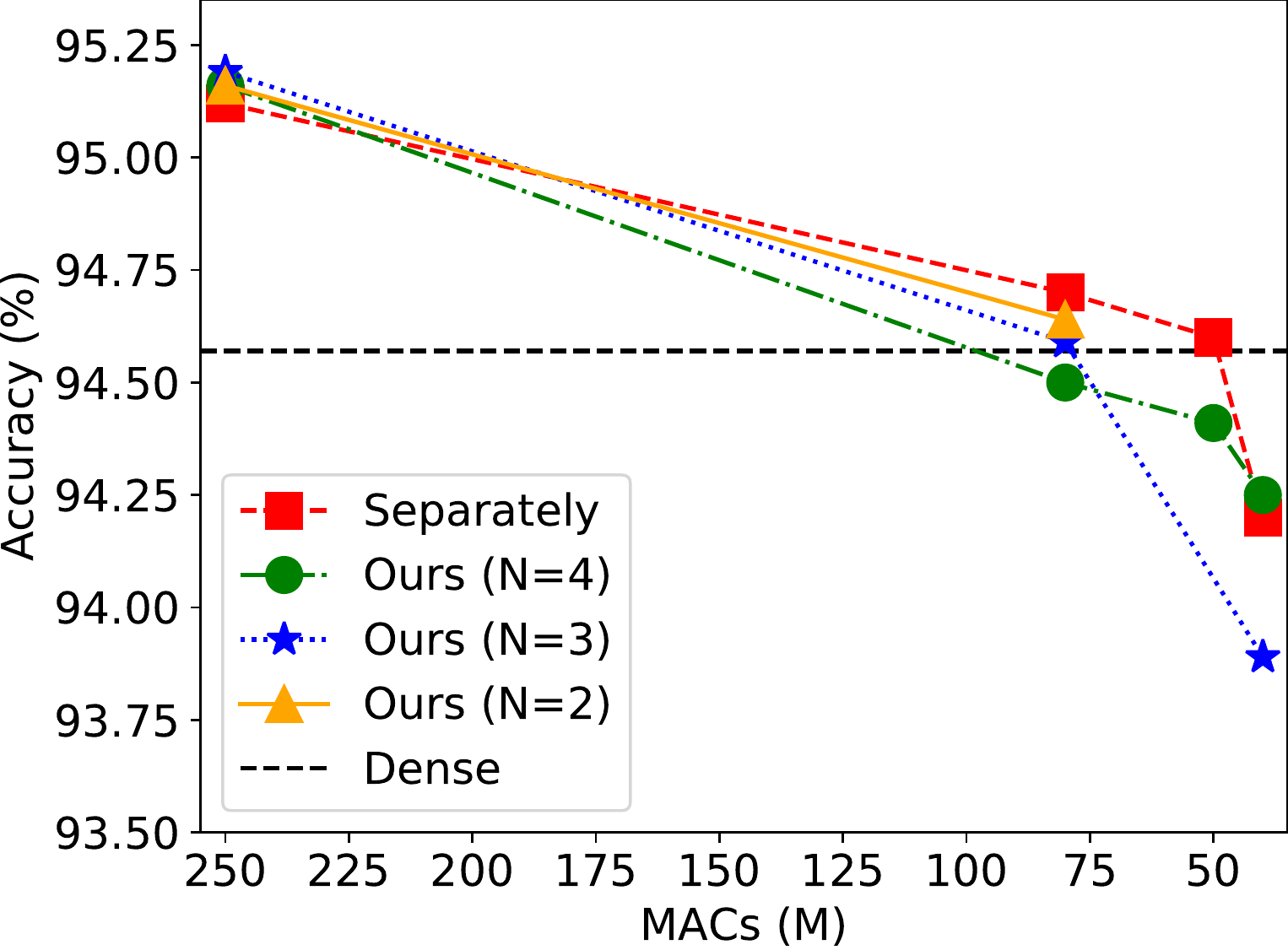}  
     \caption{Performance of All-in-One with different switch number $N$ on CIFAR-10 using Resnet-18.}
    \label{fig:various_N}  
    \vspace{-0.15in}
\end{figure}

We compare the accuracy performance of various $N$ ($N=2, 3, \text{or} \  4$) in Figure~\ref{fig:various_N}. The model architecture is Resnet-18
 and the accuracy is tested on CIFAR-10.  Pattern-based pruning is applied. We test the accuracy of the models with 250M and 80M MACs for $N=2$, models with 250M, 80M, and 40M MACs for $N=3$,  and the models with 250M, 80M, 55M, and 40M MACs for $N=4$, respectively. We also show the accuracy of the single sparse models which are separately trained for   each target MACs.  
 
As shown in Figure~\ref{fig:various_N}, generally a lower target MACs leads to a lower accuracy. Different $N$ values does not lead to significant changes on the accuracy. For example, the max accuracy difference of various $N$ and separately trained models at each target MACs is no larger than 0.5\%. In some cases, our method can achieve even higher accuracy than the separately trained models (such as $N=4$ for the model with 40M MACs).   Moreover, though $N$ increases, the memory consumption of our method  does not increase as we only need to store one model and one soft mask for various $N$.

\section{Conclusion}
We propose All-in-One, a highly representative pruning framework to work with dynamic power management using DVFS. Extensive experiments and real-world edge device evaluations show that All-in-One maintains high accuracy for each switch with low memory cost, and provide stable inference speed performance. 
It indicates that our work can deal with the limited hardware resource and unstable energy simultaneously on the widely adopted edge device. 

\section{Acknowledgement}
This work is partly supported by the Army Research Office/Army Research Laboratory via grant W911-NF-20-1-0167 to Northeastern University, National Science Foundation CCF-1937500, and CNS-1909172.

\bibliographystyle{ACM-Reference-Format}
\bibliography{sample-base}

\appendix

\end{document}